\pdfoutput=1
\documentclass[11pt]{article}

\usepackage{acl}

\usepackage{times}
\usepackage{latexsym}
\usepackage{amsmath}
\usepackage{enumitem}
\usepackage{subfig} % 用于c创建子图
\usepackage{cleveref} % 用于引用图表
\usepackage{booktabs} % 用于创建漂亮的表格线
\usepackage{array} % 用于在表格中使用 p{} 来控制列宽
\usepackage{graphicx} % 用于缩放表格
\usepackage{multirow} % 用于设置表格多行
\usepackage{listings} % 用于JSON自动换行
\usepackage[table,xcdraw]{xcolor} % 用于设置表格底色 & 文本颜色
\usepackage[most]{tcolorbox}
\usepackage{color}
\definecolor{darkgreen}{RGB}{164,226,198}
\definecolor{newblue}{RGB}{39,117,182}
\definecolor{myblue}{RGB}{215,238,247}
\newcommand{\emoji}{\raisebox{-0.4ex}{\includegraphics[height=1.2em]{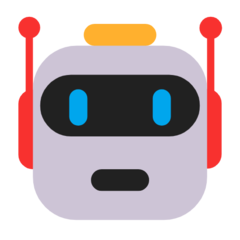}}}
\usepackage[T1]{fontenc}
\usepackage[utf8]{inputenc}

\usepackage{microtype}

\usepackage{inconsolata}

\usepackage{graphicx}

\title{Does Visual Grounding Enhance the Understanding of Embodied Knowledge in Large Language Models?}

\author{
  Zhihui Yang\textsuperscript{1} \quad
  Yupei Wang\textsuperscript{1} \quad
  Kaijie Mo\textsuperscript{1} \quad
  Zhe Zhao\textsuperscript{2} \quad
  Renfen Hu\textsuperscript{1}\thanks{Corresponding author.}
  \\
  \textsuperscript{1}Beijing Normal University \hspace{1em}
  \textsuperscript{2}Tencent AI Lab \\
  \texttt{\{yangzhihui, wangyupei, mokaijie, irishu\}@mail.bnu.edu.cn} \\
  \texttt{nlpzhezhao@tencent.com}
}

\begin{document}
\maketitle
\begin{abstract}

% Embodied knowledge is crucial for AI models to bridge the gap between the digital and physical world. 
Despite significant progress in multimodal language models (LMs), it remains unclear whether visual grounding enhances their understanding of embodied knowledge compared to text-only models. To address this question, we propose a novel embodied knowledge understanding benchmark based on the perceptual theory from psychology, encompassing visual, auditory, tactile, gustatory, olfactory external senses, and interoception. 
The benchmark assesses the models' perceptual abilities across different sensory modalities through vector comparison and question-answering tasks with over 1,700 questions. 
By comparing 30 state-of-the-art LMs, we surprisingly find that vision-language models (VLMs) do not outperform text-only models in either task. Moreover, the models perform significantly worse in the visual dimension compared to other sensory dimensions. 
Further analysis reveals that the vector representations are easily influenced by word form and frequency, and the models struggle to answer questions involving spatial perception and reasoning. Our findings underscore the need for more effective integration of embodied knowledge in LMs to enhance their understanding of the physical world\footnote{Our dataset and code are available at \url{https://github.com/ererdewubudesi/embodied_knowledge_2025}.}.

\end{abstract}

% 公式
% This an example cross-reference to Equation~\ref{eq:example}.
% \begin{equation}
%   \label{eq:example}
%   A = \pi r^2
% \end{equation}

% 图
% \begin{figure*}[t]
%   \includegraphics[width=0.48\linewidth]{example-image-a} \hfill
%   \includegraphics[width=0.48\linewidth]{example-image-b}
%   \caption {A minimal working example to demonstrate how to place
%     two images side-by-side.}
% \end{figure*}

% 引用
% \citep{BI2021883}
% \citealp{BI2021883}
% \citet{BI2021883}
% \citeyearpar{BI2021883}
% \citeposs{BI2021883}
    
\section{Introduction}

Embodied knowledge is acquired through experience and contextualized in relation to the body \citep{castree2013dictionary}. It is inherently sensory, encompassing sights, sounds, smells, touch, and taste \citep{ellingson2008embodied}. 
Humans rely on both sensory experiences and language experience to represent knowledge \citep{vinaya2024context, jones2024does, jones-etal-2024-multimodal, Andrews2014ReconcilingEA, Andrews2009IntegratingEA, BI2021883, gunther2019vector, davis2021building, kim2019knowledge}.
Similarly, embodied knowledge is crucial for AI to bridge the gap between the digital and physical world \citep{lungarella200750, liu2024aligning}.
% Humans heavily rely on sensory experiences to represent knowledge \citep{BI2021883}. 

% 去掉models
% Humans heavily rely on sensory experiences to represent knowledge \citep{BI2021883}, and similarly, embodied knowledge is crucial for AI models to bridge the gap between the digital and physical world \citep{lungarella200750, liu2024aligning}.

However, LMs have traditionally relied on training on massive textual data to understand the world, following the Distributional Hypothesis \citep{harris1954distributional}. Although they can memorize knowledge from the data and learn to utilize statistical patterns to demonstrate language understanding and generation abilities, they often make mistakes on questions related to the real world due to a lack of grounding~\citep{bender-koller-2020-climbing, merrill2021provable}. As shown in Table~\ref{tab:Bert's Masked Word Prediction Failures}, the text-only model BERT incorrectly identifies hummingbirds as the largest birds in the world and is confused about the relative sizes of objects, indicating that the model struggles to differentiate between antonyms with distinct sensory contrasts, such as \textit{small-big}. Therefore, increasing work has advocated for grounded language learning through the integration of perceptual information and interaction with the physical and social world~\citep{bisk-etal-2020-experience, bender2020climbing, ma-etal-2023-world, shi-etal-2025-learning}.

% 结尾只举一个例子，压缩篇幅。
% However, LMs have traditionally relied on training on massive textual data to understand the world, following the distributional hypothesis \citep{harris1954distributional}. Although they can memorize knowledge from the data and learn to utilize statistical patterns to demonstrate language understanding and generation abilities, they often make mistakes on questions related to the real world due to a lack of grounding~\citep{bender-koller-2020-climbing, merrill2021provable}. As shown in Table~\ref{tab:Bert's Masked Word Prediction Failures}, the text-only model BERT incorrectly identifies hummingbirds as the largest birds in the world and is confused about the relative sizes of objects, indicating that the model struggles to differentiate between antonyms with distinct sensory contrasts, such as \textit{small-big} and \textit{hot-cold}.

\begin{table}[t]
    \resizebox{0.48\textwidth}{!}{ % 调整宽度，高度自动适应
    \begin{tabular}{ll}
        \toprule
        \textbf{Prompt \& Predicted Words} \\
        \midrule
        \parbox[t]{8cm}{Hummingbirds are the [MASK] birds in the world.} \\ \textcolor{red}{largest} (0.47), \textcolor[rgb]{0,0.5,0}{smallest} (0.13), fastest (0.11), only (0.04)\\
        \midrule
        \parbox[t]{8cm}{The trophy doesn't fit into the brown suitcase because the trophy is too [MASK].} \\ \textcolor{red}{small} (0.25), \textcolor[rgb]{0,0.5,0}{big} (0.23), heavy (0.17), large (0.17)\\
        \midrule
        \parbox[t]{8cm}{The trophy doesn't fit into the brown suitcase because the suitcase is too [MASK].} \\ \textcolor{red}{big} (0.28), \textcolor[rgb]{0,0.5,0}{small} (0.22), large (0.17), heavy (0.16)\\
        \bottomrule
    \end{tabular}
  }
    \vspace{-1mm}
    \caption{Word prediction examples by \texttt{BERT-base}.}
    \vspace{-5mm}
    \label{tab:Bert's Masked Word Prediction Failures}
\end{table}

\begin{figure*}[t]
    \centering
    \includegraphics[width=1\linewidth]{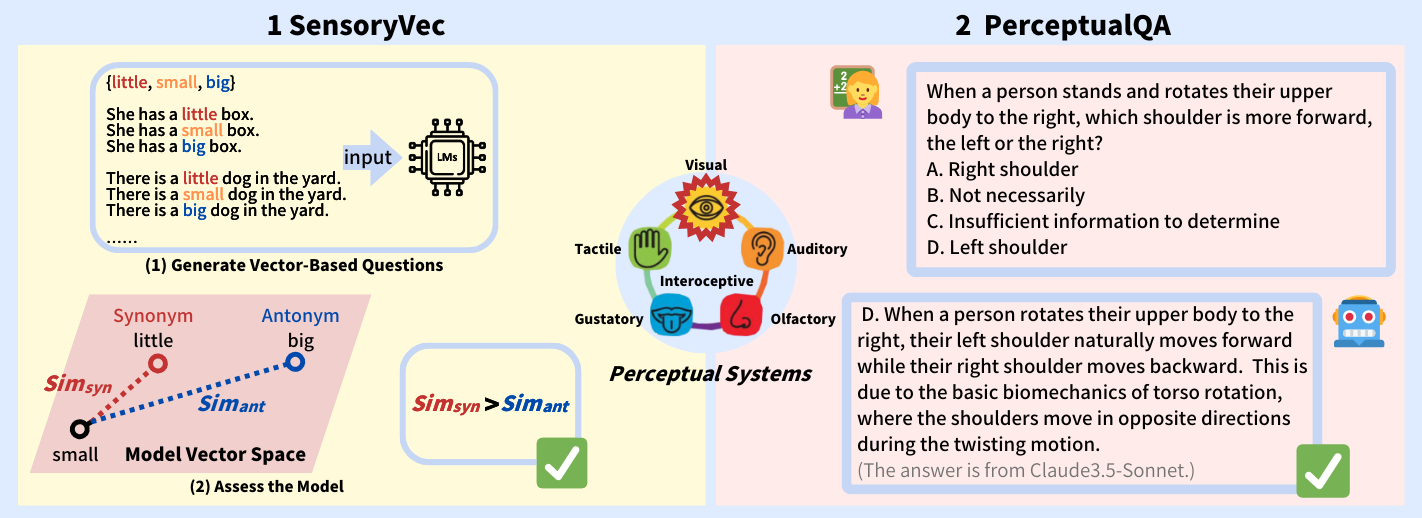}
    \vspace{-5mm}
    \caption{The two tasks in our embodied knowledge understanding benchmark.}
    \label{fig:Method}
    \vspace{-5mm}
\end{figure*}

In recent years, large language models (LLMs) have demonstrated significantly enhanced intelligence~\citep{du2022surveyvisionlanguagepretrainedmodels}. Notably, multimodal LLMs that associate natural language with visual information exhibit robust visual question-answering capabilities~\citep{yin2023survey}.
% 没增加马的论文
This observation naturally raises an intriguing question: \textbf{does visual grounding improve the models' understanding of embodied knowledge}, thereby enabling them to better perceive and comprehend the real world?

% 增加马的论文，但感觉必要不大
% Despite the exciting performance of these models on a variety of downstream tasks, it remains unclear whether these models can truly understand language with their grounded meanings~\citep{ma-etal-2023-world}, which naturally raises an intriguing question: \textbf{does visual grounding improve the models' understanding of embodied knowledge}, thereby enabling them to better perceive and comprehend the real world?

% 可视篇幅需求压缩（待定）
To address this question, we propose an embodied knowledge understanding benchmark consisting of two tasks: \texttt{SensoryVec} and \texttt{PerceptualQA}, which evaluate the model's capability to represent sensory information in vector form and answer perception-related questions, respectively.
As shown in Figure~\ref{fig:Method}, both tasks are systematically designed based on the perceptual system framework from psychology~\citep{gerrig2015psychology}, encompassing experiential knowledge from visual, auditory, tactile, gustatory, and olfactory modalities that complement language-derived representations~\citep{BI2021883}.
In \texttt{SensoryVec}, we concentrate on the five aforementioned external sensory modalities and interoception, constructing a sensory adjective lexicon with 349 word triples and 1,047 sentences reflecting natural contexts. By comparing the cosine similarities between sensory words and their synonyms versus antonyms, we assess the model's vector representations.
For \texttt{PerceptualQA}, we construct a QA dataset comprising 9 tasks and 1,400 questions across five external sensory modalities. Considering the complexity of the visual modality~\citep{gerrig2015psychology}, we design 5 visual subtasks: color attributes, colors in nature, geometry and transformations, symbols, and body.

% 可视篇幅需求压缩（待定）
We evaluate a wide range of state-of-the-art LMs, specifically selecting VLMs and text-only LMs with multiple comparable pairs, such as Qwen-VL and Qwen, LLaVA-Mistral and Mistral, where the former is directly built upon the latter.
We have three main findings:
(1) Current models exhibit suboptimal performance in understanding embodied knowledge, with the best-performing models achieving only around 70\% accuracy on both tasks, far below human performance.
(2) VLMs initialized with text model parameters demonstrate no clear advantage over their language model counterparts on either task.
(3) Nearly all models perform significantly worse on visual tasks compared to other modalities, particularly on questions involving spatial understanding and reasoning in categories such as symbols, geometry, and body.
These results suggest that multimodal learning has not yet fully capitalized on its potential to capture and leverage embodied knowledge. As embodied AI research actively integrates LLMs and vision large language models (VLLMs) into decision-making~\cite{liu2024aligning}, the models' limitations in understanding and reasoning about embodied knowledge may become a significant obstacle or concern.
In summary, this paper presents a systematic, comprehensive benchmark for evaluating models' embodied knowledge across different modalities, which is valuable for model analysis and diagnosis. Moreover, the resource construction approach outlined in this paper can serve as a reference for more comprehensive model evaluation.

\section{Related Work}

% \subsection{Vector Representation}
\subsection{Comparing VLMs and text-only LMs}
The emergence of multimodal LMs has sparked interest in the differences between VLMs and text-only LMs. A series of studies have compared these two types of models based on vector representations, involving tasks such as word similarity, semantic probing, and measuring correlations between vectors and fMRI data.

\citet{pezzelle-etal-2021-word} investigated the vector representation differences between multimodal models (e.g. ViLBERT and VisualBERT) and the text-only model BERT on existing word similarity benchmarks. They discovered that multimodal models surpass BERT on concrete words.
% \citet{pezzelle-etal-2021-word} investigated the vector representation differences between multimodal models (LXMERT, UNITER, ViLBERT, VisualBERT, and Vokenization) and the text-only model BERT on existing word similarity benchmarks, such as WordSim353~\citep{finkelstein2001placing} and SimLex999~\citep{hill2015simlex}. They discovered that multimodal models surpass BERT on concrete word.
% 详细：RG65~\citep{rubenstein1965contextual}, WordSim353~\citep{finkelstein2001placing}, SimLex999~\citep{hill2015simlex}, MEN~\citep{bruni2014multimodal} and SimVerb3500~\citep{faruqui2016problems}
\citet{yun-etal-2021-vision-language} compared the embedding differences between VLMs (VideoBERT and VisualBERT) and their text-only variants using a series of probing tasks covering physical commonsense QA, coreference resolution, semantic role labeling, and adjective-noun composition.
They found that the multimodal models fail to significantly outperform the text-only variants.
\citet{bavaresco2024modellingmultimodalintegrationhuman} compared word embedding alignment to brain activity using the fMRI dataset by \citet{pereira2018toward}. Representational similarity analysis revealed substantial differences among VLMs, with some exhibiting higher brain response correlations than unimodal models.

\citet{bylinina-etal-2023-leverage} further explored the factors influencing differences in word vector similarities. They calculated word similarity data for 13,000 word pairs based on multimodal LMs (CLIP, OpenCLIP, Multilingual CLIP) and text-only models (FastText, mBERT, XLM-RoBerta). Using 46 semantic features, they conducted regression analysis to predict vector similarity differences. The results showed that concreteness and taxonomic features from WordNet were significant predictors. However, a considerable portion of the embedding space differences remained unexplained.

In summary, existing research has investigated the differences between multimodal LMs and text-only LMs. However, these studies focused only on vector representations and relied on existing evaluation datasets, which prevented them from comprehensively and specifically diagnosing differences between the two types of models, leading to somewhat contradictory conclusions~\cite{pezzelle-etal-2021-word, yun-etal-2021-vision-language, bavaresco2024modellingmultimodalintegrationhuman}. Moreover, these works mainly focused on early multimodal and text-only LMs and have not yet considered the perceptual capabilities of LLMs.

\subsection{Multimodal reasoning benchmarks}
Numerous visual reasoning benchmarks, such as CLEVR~\cite{Johnson2016CLEVRAD}, MMMU~\cite{10656299}, and MMBench~\cite{Liu2023MMBenchIY}, have been proposed to evaluate VLMs. These benchmarks typically emphasize reasoning over domain knowledge, object attributes and relationships, as well as scene understanding and prediction. Structured as visual question answering tasks, their reliance on visual input at inference time makes it difficult to directly compare the performance of text-only and vision-language models.

\subsection{Embodied AI}

Embodied AI is essential for achieving Artificial General Intelligence (AGI) and connecting the digital and physical worlds~\citep{liu2024aligning}. For tasks such as 3D Visual Grounding and Embodied Question Answering, traditional methods often struggle with complex queries and require extensive labeled data. 
Recent studies~\citep{yang2024llm, yuan2024visual, majumdar2024openeqa, patel2024embodied} show that (multimodal) LLMs can effectively handle diverse queries while reducing data dependencies, often achieving strong performance even without task-specific fine-tuning.

% 压缩篇幅
% Embodied AI plays a crucial role in achieving Artificial General Intelligence (AGI) and enables applications that bridge the digital and physical realms (Liu et al., 2024). 
% Multimodal LLMs have been introduced to embodied AI research due to their exceptional understanding and reasoning capabilities. Their impact is evident in tasks like 3D Visual Grounding and Embodied Question Answering. Traditional approaches to these tasks often struggle with complex queries and rely heavily on extensive labeled data. Recent research~\citep{yang2024llm, yuan2024visual, majumdar2024openeqa, patel2024embodied} has demonstrated that incorporating LLMs and multimodal LLMs can effectively handle diverse queries while reducing data dependencies, often achieving strong performance even without task-specific fine-tuning.

Meanwhile, embodied data has been leveraged to enhance LLM capabilities. Recent studies~\cite{yang2024binding, fu2024touch, yu2024octopi} have integrated tactile data with visual and linguistic modalities through contrastive learning and fine-tuning, enabling LLMs to better understand cross-modal embodied knowledge.

While multimodal LLMs are expected to serve as the brain of embodied agents~\citep{liu2024aligning}, bridging the gap to human-level performance remains challenging. 
Real-world applications such as autonomous driving and human-computer interaction require sophisticated object understanding, spatial reasoning, and geometric inference capabilities. Continuous assessment and improvement in these areas is essential for robust and generalizable embodied AI systems.

\section{Method}
This paper proposes an embodied knowledge understanding benchmark consisting of two tasks: \texttt{SensoryVec} and \texttt{PerceptualQA}, which evaluate the model's capability to represent sensory information in vector form and answer perception-related questions, respectively.

\begin{table*}[t]
    \centering
    \resizebox{1\textwidth}{!}{ % 调整宽度，高度自动适应
    \begin{tabular}{ll}
        \toprule
        \textbf{Task} & \textbf{Question Perspectives} \\
        \hline
        Visual-Color Attributes & Hue, brightness, saturation \\
        \hline
        Visual-Colors in Nature & Hue, saturation, warm and cool tones \\
        \hline
        Visual-Geometry and Transformations & Quantity, shape, direction, size relation, position relation \\
        \hline
        Visual-Symbols & Quantity, shape, direction, size relation, position relation \\
        \hline
        Visual-Body & Shape, direction, distance, speed, position relation \\
        \hline
        Auditory & Volume, pitch \\
        \hline
        Tactile & Smoothness, elasticity, hardness, weight, thickness, stickiness, temperature \\
        \hline
        Gustatory & Sourness, sweetness, bitterness, saltiness, spiciness \\
        \hline
        Olfactory & Fragrance, stench, specific odors \\
        \bottomrule
    \end{tabular}
  }
    \vspace{-1mm}
    \caption{Overview of subtasks in \texttt{PerceptualQA}.}
    \label{tab:benchmark_categories}
    \vspace{-2mm}
\end{table*}

% \begin{table*}[t]
%     \centering
%     % \renewcommand{\arraystretch}{1.2}
%     \resizebox{1\textwidth}{!}{ % 调整宽度，高度自动适应
%     \begin{tabular}{lcl}
%         \toprule
%         \textbf{Task} & \textbf{Number} & \textbf{Question Perspectives} \\
%         \hline
%         Visual-Color Attributes & 200 & Hue, brightness, saturation \\
%         \hline
%         Visual-Colors in Nature & 200 & Hue, saturation, warm and cool tones \\
%         \hline
%         Visual-Geometry and Transformations & 200 & Quantity, shape, direction, size relation, position relation \\
%         \hline
%         Visual-Symbols & 200 & Quantity, shape, direction, size relation, position relation \\
%         \hline
%         Visual-Body & 200 & Shape, direction, distance, speed, position relation \\
%         \hline
%         Auditory & 100 & Volume, pitch \\
%         \hline
%         Tactile & 100 & Smoothness, elasticity, hardness, weight, thickness, stickiness, temperature \\
%         \hline
%         Gustatory & 100 & Sourness, sweetness, bitterness, saltiness, spiciness \\
%         \hline
%         Olfactory & 100 & Fragrance, stench, specific odors \\
%         \bottomrule
%     \end{tabular}
%   }
%     \vspace{-2mm}
%     \caption{Overview of subtasks in \texttt{PerceptualQA}}
%     \label{tab:benchmark_categories}
%     \vspace{-2mm}
% \end{table*}

\begin{table*}[t]
    \centering
    \resizebox{1\textwidth}{!}{ % 调整宽度，高度自动适应
    \begin{tabular}{ll}
        \toprule
        \textbf{Task} & \textbf{Question} \\
        \midrule
        Visual-Color Attributes & Compared to red, what is the hue of burgundy? \\ 
        & A. More orangish \quad \textbf{B. More purplish} \quad C. More bluish \quad D. More greenish \\
        \hline
        Visual-Colors in Nature & Which of these foods has the highest color saturation? \\ 
        & A. Jellyfish \quad B. Raw eel \quad C. Raw catfish \quad \textbf{D. Cooked shrimp} \\
        \hline
        Visual-Geometry and Transformations & When a rectangle is folded so that its top and bottom edges coincide, what shape is formed? \\ 
        & \textbf{A. Rectangle} \quad B. Isosceles triangle \quad C. Rhombus \quad D. Square \\
        \hline
        Visual-Symbols & When the number [7] is rotated 45 degrees counterclockwise, which direction does its opening face? \\ 
        & A. Left \quad B. Up \quad \textbf{C. Down} \quad D. Right \\
        \hline
        Visual-Body & When standing with hands behind the back, are the elbows higher or lower than the hips? \\ 
        & \textbf{A. Higher} \quad B. Lower \quad C. At the same height \quad D. It varies \\
        \hline
        Auditory & Which sound is usually louder: the sound of frying fish or the sound of steaming fish? \\ 
        & A. Steaming fish \quad \textbf{B. Frying fish} \quad C. Almost the same \quad D. It depends \\
        \hline
        Tactile & Which fruit has the roughest skin? \\ 
        & A. Persimmon \quad B. Banana \quad C. Pear \quad \textbf{D. Pineapple} \\
        \hline
        Gustatory & Which food is least likely to have a bitter taste? \\ 
        & A. Coffee \quad B. Bitter gourd \quad \textbf{C. Corn} \quad D. Green tea \\
        \hline
        Olfactory & Which of the following things has the most distinct fragrance? \\ 
        & A. Eggplant \quad \textbf{B. Rose tea} \quad C. Carrot \quad D. Potato \\
        \bottomrule
    \end{tabular}
    }
    \vspace{-1mm}
    \caption{Sample questions from \texttt{PerceptualQA} (correct answers highlighted in \textbf{bold}).}
    \label{tab:example_of_tasks}
    \vspace{-5mm}
\end{table*}

\subsection{SensoryVec Task} 
\label{sec:SensoryVec Task}

% 方法流程
% 新增任务设计动机的详细解释附录指引
Given LMs face difficulty in distinguishing antonyms with sensory contrasts (see examples in Table~\ref{tab:Bert's Masked Word Prediction Failures}), we evaluated the models' vector representations specifically for sensory adjectives. By comparing the similarities between each word and its synonym versus its antonym, we examined whether models assign higher similarity to the synonymous term, as illustrated in Figure~\ref{fig:Method}. A detailed motivation for designing \texttt{SensoryVec} task can be seen in Appendix~\ref{sec:Motivation Behind SensoryVec Task Design}.
% \footnote{A detailed motivation for designing \texttt{SensoryVec} task can be seen in Appendix~\ref{sec:Motivation Behind SensoryVec Task Design}.}

First, we collected candidate sensory adjectives from three high-quality datasets~\citep{lievers2015synaesthesia, lynott2009modality, Lynott2019TheLS}.
Referring to the perceptual system framework from psychology~\cite{gerrig2015psychology}, we selected sensory words related to visual, auditory, tactile, gustatory, olfactory, and interoceptive senses, and annotate their attributes. To ensure representativeness, we retained words with sensory ratings above 4 from the datasets constructed by~\citet{lynott2009modality} and~\citet{Lynott2019TheLS}. Subsequently, we reviewed and cleaned the data (e.g., by removing duplicate words), as detailed in Appendix~\ref{sec:Word Selection Criteria}.

% 原啰嗦版，去掉三个原则，并精炼数据生成过程：Next, synonyms and antonyms are identified with reference to WordNet \citep{fellbaum1998wordnet} and a thesaurus \citep{random2001random}. For words without strict antonyms, such as color terms, semantically distant counterparts are used. In this way, we construct sensory triples like \textit{small-little-big}. Furthermore, we curate three natural contextual sentences for each triple. These sentences are generated through collaboration between LLMs and human annotation (details in Appendix~\ref{sec:Sentence Data Construction Details}), adhering to the following principles: (1) grammaticality, (2) semantic appropriateness, and (3) contextual alignment, matching polysemous words' meanings to their primary sensory category. Finally, we obtain 349 groups of sensory adjectives and 1047 sentences, systematically covering a wide range of attributes across different sensory dimensions:
Next, synonyms and antonyms are identified with reference to WordNet~\citep{fellbaum1998wordnet} and a thesaurus~\citep{random2001random}. For words without strict antonyms, such as color terms, semantically distant counterparts are used. In this way, we constructed sensory triples like \textit{small-little-big}. Furthermore, we curated three natural contextual sentences for each triple, generated by LLMs and then refined through human review (details in Appendix~\ref{sec:Sentence Data Construction Details}).
Finally, we obtained 349 groups of sensory adjectives and 1047 sentences, systematically covering a wide range of attributes across different sensory dimensions:
\begin{itemize}[noitemsep, topsep=0pt]
    \item \textbf{Visual}: color, shape, size, depth, distance, orientation, displacement, speed, etc.
    \item \textbf{Auditory}: pitch, loudness, rhythm, etc.
    \item \textbf{Tactile}: texture, cold, warm, etc.
    \item \textbf{Gustatory}: sour, sweet, bitter, spicy, salty, specific food flavors, etc.
    \item \textbf{Olfactory}: fragrant, odorous, pungent, etc.
    \item \textbf{Interoceptive}: hunger, satiety, etc.
\end{itemize}

In the evaluation, for contextualized models, we use sentences as input and obtain average token vectors from the last hidden layer, then compute the mean vector across the three sentences. For other models (such as Word2vec and GloVe), we directly utilize standalone word vectors.

\subsection{PerceptualQA Task} 

Considering that question-answering is the most natural interaction mode for LLMs, we introduce the \texttt{PerceptualQA} task to comprehensively evaluate a model's embodied knowledge. Consistent with the \texttt{SensoryVec} task, we design questions referring to perceptual systems from psychological theory \citep{gerrig2015psychology}, covering visual, auditory, tactile, gustatory, and olfactory modalities. Given the complexity of vision and the emphasis on visual training in most existing multimodal models, we design five subtasks for the visual modality: color attributes, colors in nature, geometry and transformations, symbols, and body. 

% In each subtask, questions are designed from multiple angles. 
As shown in Table~\ref{tab:benchmark_categories}, questions for each modality are designed from multiple perspectives. They are intended to be easily answerable by humans through embodied imagination and reasoning.\footnote{A detailed motivation for designing \texttt{PerceptualQA} task can be seen in Appendix~\ref{sec:Motivation Behind PerceptualQA Task Design}.}
For geometry, symbol, and body-related questions, we consider not only the original shapes of objects but also their transformed states (e.g., after rotation). Table~\ref{tab:example_of_tasks} presents examples for each task. The full question frameworks are provided in Table~\ref{tab:question_frameworks_resized} (see Appendix~\ref{sec:Question Frameworks}).

% As shown in Table \ref{tab:benchmark_categories}, each modality is approached from multiple angles. For instance, regarding color, we not only focus on basic attributes such as hue, brightness, and saturation but also inquire about color groupings and comparisons in animals and plants. For geometry, symbol, and body-related questions, we introduce transformations such as rotation. These questions are easily answerable by humans through embodied imagination and reasoning. Table~\ref{tab:example_of_tasks} provides examples of each task.
% We prompted an LLM (Claude-3.5-Sonnet) as a brainstorming aid to generate candidate questions to support human annotators. The majority of \texttt{PerceptualQA} items are manually crafted to meet rigorous design criteria, as most model-generated questions are discarded due to lacking a correct or unique answer, or being simply infeasible to answer. A small subset of questions originating from the model are manually screened and refined according to the same criteria.
We used an LLM (Claude-3.5-Sonnet) as a brainstorming tool to generate candidate questions for human annotators. Each question has four options with one correct answer. 
Most \texttt{PerceptualQA} items were manually filtered; the remaining items were further revised by annotators according to rigorous design criteria.
Ultimately, the PerceptualQA task consists of 1,400 multiple-choice questions, with 200 questions per visual subtask and 100 questions per other perceptual understanding task. See details of dataset construction in Appendix~\ref{sec:QA Construction Details}.

% We prompte an LLM to generate candidate questions based on the above framework. These questions are manually screened and refined according to predefined guidelines. Each question has four options with one correct answer. Ultimately, the \texttt{PerceptualQA} task consists of 1,400 multiple-choice questions, with 200 questions per visual subtask and 100 questions per other perceptual understanding task. See details of dataset construction in Appendix~\ref{sec:QA Construction Details}.

\section{Experiments}

\subsection{Models and Settings}
% 选择依据和范围
We systematically evaluated an extensive range of state-of-the-art VLMs and text-only models, spanning diverse architectures, model sizes, and availability (open-source vs. closed-source). Models assessed include CLIP, Word2Vec, GloVe, BERT, Mistral, Vicuna, LLaMA, Gemma, Qwen, LLaVA, GPT, Gemini, and Claude series. See Appendix~\ref{sec:model} for details of models and parameters.

% 可比模型
The selected models include 6 groups of comparable model pairs, where a vision-language model is built upon a text-only language model, enabling direct exploration of the impact of visual grounding. These include VisualBERT \& BERT, LLaVA-1.6-Vicuna-7B \& Vicuna-7B, LLaVA-1.6-Mistral-7B \& Mistral-7B, Qwen-VL-Chat \& Qwen-7B, Qwen2-VL-7B-Instruct \& Qwen2-7B, and Qwen2-VL-72B-Instruct \& Qwen2-72B~\citep{Li2019VisualBertAS, liu2024visual, liu2024improved, Qwen-VL, wang2024qwen2vlenhancingvisionlanguagemodels}.

% 评价指标
For \texttt{SensoryVec}, we evaluated the accuracy based on cosine similarity judgments for each triple.\footnote{Triples containing words that are not in Word2Vec and GloVe vocabulary are excluded from total count.}
For \texttt{PerceptualQA}, we reported average accuracy across two trials for reliability.\footnote{Evaluation prompts are in Appendix~\ref{sec:model_PerceptualQA Task}, with correct answer distributions shown in Table~\ref{tab:distribution_correct_answers}.}

\subsection{Results} 
\subsubsection{SensoryVec Task}

\begin{table}[t]
  \centering
  \resizebox{0.48\textwidth}{!}{
  \begin{tabular}{lccc}
    \toprule
    \textbf{Model} & \textbf{All Sensory} & \textbf{Visual} & \textbf{Non-Visual} \\
    \midrule
    \midrule
    Word2Vec & 67.64 & 65.00 & 71.33 \\
    GloVe & 62.50 & 57.81 & 69.12 \\
    \midrule
    \rowcolor{gray!30} BERT & \textbf{72.21} & 70.44 & \textbf{74.66} \\
    \rowcolor{gray!30} \underline{VisualBERT} & 64.18 & 65.52 & 62.33 \\
    \midrule
    GPT-2 & 50.43 & 47.78 & 54.11 \\
    \rowcolor{gray!15} Qwen-7B & 61.32 & 54.68 & 70.55 \\
    \rowcolor{gray!15} \underline{Qwen-VL} & 58.74 & 53.69 & 65.75 \\
    \rowcolor{gray!15} \underline{Qwen-VL-Chat} & 61.60 & 55.17 & 70.55 \\
    \rowcolor{gray!30} Qwen2-7B & 63.32 & 58.62 & 69.86 \\
    \rowcolor{gray!30} Qwen2-7B-Instruct & 66.19 & 61.58 & 72.60 \\
    \rowcolor{gray!30} \underline{Qwen2-VL-7B-Instruct} & 63.04 & 58.62 & 69.18 \\
    \rowcolor{gray!15} Mistral-7B & 67.05 & 63.55 & 71.92 \\
    \rowcolor{gray!15} \underline{LLaVA1.6-Mistral-7B} & 66.76 & 65.02 & 69.18 \\
    \rowcolor{gray!30} Vicuna-7B & 57.59 & 58.62 & 56.16 \\
    \rowcolor{gray!30} \underline{LLaVA1.6-Vicuna-7B} & 58.45 & 59.61 & 56.85 \\
    \midrule
    \underline{CLIP} & 71.06 & \textbf{75.37} & 65.07 \\
    \bottomrule
  \end{tabular}
  }
  \vspace{-2mm}
  \caption{Results on \texttt{SensoryVec} for static embeddings, bidirectional, generative, and contrastive learning models (top to bottom). VLMs are \underline{underlined}. Comparable models within each group share the same background color, with the LMs above the VLMs. \textbf{Bolded} values denote the highest accuracy per sensory modality.}
  \label{tab:accuracy_vector-based}
  \vspace{-5mm}
\end{table}

Table~\ref{tab:accuracy_vector-based} demonstrates the results on \texttt{SensoryVec}. All models perform suboptimally on this task, with accuracy ranging from 50\% to 70\%. Moreover, VLMs perform comparably or worse than their corresponding text-only models, consistent with \citet{yun-etal-2021-vision-language}'s findings. This reveals that perceiving sensory contrasts remains an important challenge for both text-only and multimodal models. Furthermore, across different modalities, we find that for most models, visual word representations are notably inferior to those of non-visual sensory words. Even VLLMs like Qwen2-VL-7B-Instruct and LLaVA1.6-Mistral-7B struggle to bridge this gap.
Surprisingly, the text-only language model BERT achieves the best overall accuracy at 72.21\%. This indicates that even without explicit visual input, models can capture a significant amount of sensory-related information from large-scale textual data, although substantial gaps remain. 
These results echo findings from psychological studies showing that language can convey certain aspects of perceptual knowledge. For example, previous research has demonstrated that congenitally blind individuals can acquire some knowledge about visual properties such as shape, texture, or size through linguistic descriptions and conceptual inference, despite not having direct visual experience~\citep{BI2021883, kim2019knowledge}. It is important to note that, unlike language models, blind individuals have access to other sensory modalities (e.g., touch and hearing), providing them with richer embodied experiences.
Nevertheless, the absence of sensory experience still hinders understanding of some vision-dependent attributes, such as color. In ~\citet{kim2019knowledge}, blind participants exhibited significant difficulty grouping and sorting animal colors compared to sighted controls.

% 不要：Surprisingly, the text-only language model BERT achieved the best overall accuracy at 72.21\%. This suggests that even without explicit visual input, models can capture a significant amount of sensory information from large-scale textual data, although considerable gaps persist.
% 不要：hese results align with psychological studies~\citep{kim2019knowledge} on blind individuals, who exhibit substantial agreement with sighted individuals on certain appearance dimensions (e.g., shape, texture, size, height) by inferring characteristics through reasoning (e.g., taxonomic knowledge). However, the absence of sensory experience still hinders understanding of some vision-dependent attributes, such as color. In ~\citet{kim2019knowledge}'s experiments, compared to sighted subjects, blind subjects faced significant difficulties in grouping and sorting animal colors.

Our detailed analysis of \texttt{SensoryVec} results suggests that the models’ suboptimal performance is primarily due to an over-reliance on distributional semantics during training, as well as insufficient representation of form-similar and low-frequency terms. Section~\ref{tab:Discussion} and Table~\ref{tab:Models' Vector-Based Prediction Failures} illustrate these issues with detailed examples.

\subsubsection{PerceptualQA Task}

\begin{table}[t]
  \centering
  \resizebox{0.48\textwidth}{!}{
    \begin{tabular}{lcccc}
      \toprule
      \textbf{Model} & \textbf{All Sensory} & \textbf{Visual} & \textbf{Non-Visual} \\
      \midrule
      \midrule
      Human Baseline & 86.00 & 85.20 & 88.00\\
      \midrule
      Llama3.2-3B-Instruct & 47.71 & 38.25 & 71.38 \\
      \rowcolor{gray!30} Vicuna-7B & 38.25 & 32.60 & 52.38 \\
      \rowcolor{gray!30} \underline{LLaVA1.6-Vicuna-7B} & 41.64 & 35.25 & 57.63 \\
      \rowcolor{gray!15} Mistral-7B & 42.96 & 31.40 & 71.88 \\
      \rowcolor{gray!15} \underline{LLaVA1.6-Mistral-7B} & 45.64 & 35.90 & 70.00 \\
      \rowcolor{gray!30} Qwen2-7B & 49.36 & 39.85 & 73.13 \\
      \rowcolor{gray!30} Qwen2-7B-Instruct & 47.82 & 35.80 & 77.88 \\
      \rowcolor{gray!30} \underline{Qwen2-VL-7B-Instruct} & 51.00 & 41.70 & 74.25 \\
      Llama3.1-8B & 48.54 & 39.05 & 72.25 \\
      Gemma2-9B & 51.89 & 40.65 & 80.00 \\
      Gemma2-27B & 55.39 & 44.90 & 81.63 \\
      Llama3.1-70B & 59.71 & 49.85 & 84.38 \\
      \rowcolor{gray!15}  Qwen2-72B-Instruct & 62.32 & 52.75 & 86.25 \\
      \rowcolor{gray!15} \underline{Qwen2-VL-72B-Instruct} & 63.89 & 54.45 & 87.50 \\
      Llama3.1-405B & 63.46 & 54.55 & 85.75 \\
      \midrule
      GPT-3.5 & 50.46 & 39.50 & 77.88 \\
      Qwen-Max & 68.71 & \textbf{61.05} & 87.88 \\
      \midrule
      \underline{GPT-4o-Mini} & 57.18 & 46.35 & 84.25 \\
      \underline{Gemini1.5-Flash-8B} & 54.39 & 44.55 & 79.00 \\
      \underline{Gemini1.5-Flash} & 56.07 & 45.20 & 83.25 \\
      \underline{GPT-4o} & 68.46 & 59.45 & 91.00 \\
      \underline{Gemini1.5-Pro} & 65.21 & 56.55 & 86.88 \\
      \underline{Claude3.5-Sonnet} & \textbf{69.04} & 60.00 & \textbf{91.63} \\
      \underline{Qwen-VL-Max} & 64.68 & 55.30 & 88.13 \\
      \bottomrule
    \end{tabular}
  }
  \vspace{-2mm}
  \caption{Results on \texttt{PerceptualQA} for open-source models, closed-source LLMs, and closed-source VLLMs (top to bottom). VLLMs are \underline{underlined}. Comparable models within each group share the same background color, with LLMs above VLLMs. \textbf{Bolded} values denote the highest accuracy per sensory modality.}
  \label{tab:accuracy_QA-based}
  \vspace{-5mm}
\end{table}

% For the \texttt{PerceptualQA} task, we can evaluate state-of-the-art closed-source models through question-answering. 
% However, as shown in Table~\ref{tab:accuracy_QA-based}, all models still perform poorly (see~\ref{sec:Detailed Results on PerceptualQA} for complete results). The best model, Claude3.5-Sonnet, achieves an accuracy of 69.04\%, significantly lower than human performance (86.00\%).\footnote{To provide a reference for evaluating model performance, we recruited native speakers to establish a human baseline, with detailed information available in Appendix~\ref{sec:Human Participant Information}.}

% For the \texttt{PerceptualQA} task, we  evaluate state-of-the-art closed-source models through question-answering. 
% However, as shown in Table~\ref{tab:accuracy_QA-based}, all models still perform poorly (see Appendix~\ref{sec:Detailed Results on PerceptualQA} for complete results). 
Table~\ref{tab:accuracy_QA-based} presents the results for the \texttt{PerceptualQA} task. All models continue to perform poorly (see Appendix~\ref{sec:Detailed Results on PerceptualQA} for complete results).
The best model, Claude3.5-Sonnet, achieves an accuracy of 69.04\%, significantly lower than human performance (86.00\%).\footnote{To establish a human baseline for model evaluation, we recruited seven native-speaking graduate students from the university community. Each question received two independent responses, yielding a Cohen’s Kappa of 0.69. Participants completed the tasks individually using Wenjuanxing, an online survey platform with functionality comparable to Amazon Mechanical Turk. All participants reported good health and no relevant impairments. Further details are provided in Appendix~\ref{sec:Human Participant Information}.}

Across four comparable model groups, VLLMs (LLaVA1.6-Vicuna-7B, 41.64\%; LLaVA1.6-Mistral-7B, 45.64\%; Qwen2-VL-7B-Instruct, 51.00\%; Qwen2-VL-72B-Instruct, 63.89\%) show no clear advantage over their LLM counterparts (Vicuna-7B, 38.25\%; Mistral-7B, 42.96\%; Qwen2-7B, 49.36\%; Qwen2-72B-Instruct, 62.32\%), with an average accuracy gain of only 2.32\%. This indicates that incorporating visual information does not substantially enhance model performance on these tasks. Furthermore, these findings exhibit cross-language generalizability (see Appendix~\ref{sec:Cross-Language Generalization}).

% 理论来说这两张图应当放在Discussion章节内！！！但可拿出来便于排版。
\begin{figure*}[h]
    \centering
    \subfloat{
        \includegraphics[width=0.48\textwidth]{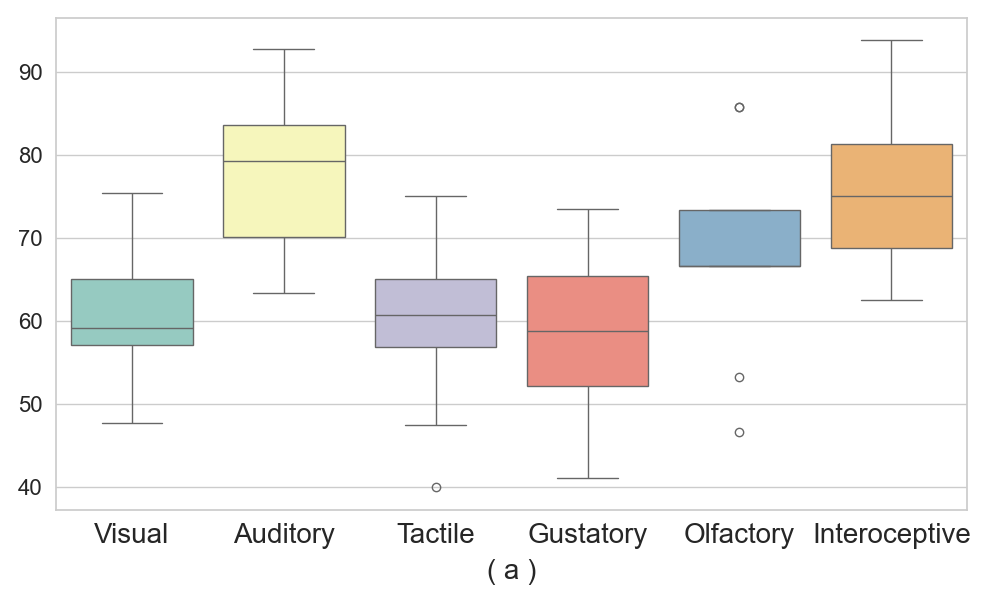}
        \label{fig:box1}
    }
    \hfill
    \subfloat{
        \includegraphics[width=0.48\textwidth]{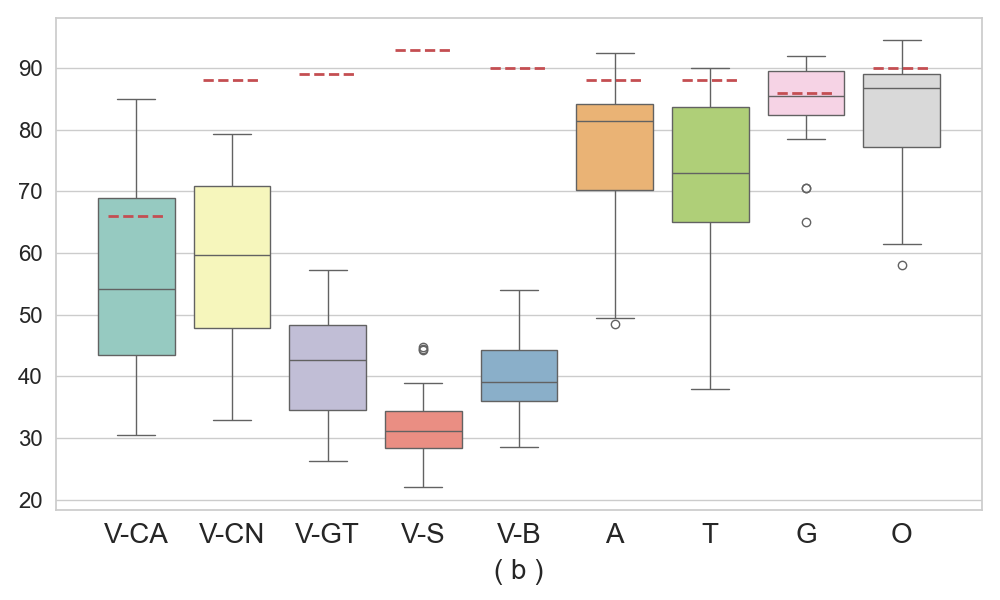}
        \label{fig:box2}
    }
    % \vspace{-2mm}
    % \vspace{-4mm}
    \caption{Box plots of accuracy on \texttt{SensoryVec} (a) and \texttt{PerceptualQA} (b) subtasks. \texttt{PerceptualQA} subtasks include Visual-Color Attributes(V-CA), Visual-Colors in Nature(V-CN), Visual-Geometry and Transformations(V-GT), Visual-Symbols(V-S), Visual-Body(V-B), Auditory(A), Tactile(T), Gustatory(G), and Olfactory(O). Red dashed lines indicate human performance.}
    % \vspace{-2mm}    
\end{figure*}

\begin{table*}[h]
\centering
\renewcommand{\arraystretch}{0.95}  % 设置行高为默认行高的 95%
\resizebox{1\textwidth}{!}{
    \begin{tabular}{lll}
    \toprule
    \textbf{Category} & \textbf{Triple (word, synonym, antonym)} & \textbf{Context} \\
    \midrule
    Tactile & (dry, waterless, wet) & The towel was completely (dry, waterless, wet) now. \\
    Tactile & (sharp, honed, blunt) & The knife is (sharp, honed, blunt). \\
    \midrule
    Gustatory & (sugarless, unsweetened, sugary) & She prefers (sugarless, unsweetened, sugary) drinks. \\
    Gustatory & (salty, brackish, flavorless) & The soup tasted (salty, brackish, flavorless). \\
    \midrule
    Visual-Shape & (concave, hollow, convex) & The sculpture's surface was (concave, hollow, convex). \\
    Visual-Shape & (loose, baggy, tight) & She prefers her jeans to be (loose, baggy, tight). \\
    Visual-Shape & (unwrinkled, smooth, wrinkled) & The surface feels (unwrinkled, smooth, wrinkled). \\
    Visual-Shape & (open, unclosed, closed) & The gate was left (open, unclosed, closed). \\
    \midrule
    Visual-Color & (black, sable, white) & The walls of the room were painted (black, sable, white). \\
    Visual-Color & (brown, chestnut, blue) & Her hair is (brown, chestnut, blue). \\
    Visual-Color & (green, verdant, red) & She wore a pair of (green, verdant, red) socks. \\
    Visual-Color & (undimmed, bright, dim) & The stars in the night sky were (undimmed, bright, dim). \\
    \bottomrule
    \end{tabular}
}
% \vspace{-2mm}
\caption{Examples of models' prediction failures on \texttt{SensoryVec}, with $Sim_{antonym} > Sim_{synonym}$.}
% \vspace{-3mm}
\vspace{-5mm}
\label{tab:Models' Vector-Based Prediction Failures}
\end{table*}
% 理论来说这两张图应当放在Discussion章节内！！！但可拿出来便于排版。

% 跨任务比较：视觉任务远不如其他任务，与人类不同
Similar to \texttt{SensoryVec}, Table~\ref{tab:accuracy_QA-based} reveals that all models perform substantially worse on visual tasks compared to other tasks. However, for humans, the difficulty gap between visual (85.20\%) and non-visual (88.00\%) tasks is minimal.  This comparison suggests that models, unlike humans, face greater difficulty in certain sensory dimensions. The best model performance on visual tasks (61.05\%) falls significantly short of human performance (85.20\%). Additionally, we analyzed the influence of potential confounding variables (see Appendix~\ref{sec:Confound variables}). In Section~\ref{tab:Discussion}, we further discuss the performances of the different subtasks.

% "while the questions themselves have comparable difficulty levels for humans" 改为"与人类相比"，同时最好去掉"rather than "那句，回应审稿人质疑：Similar to \texttt{SensoryVec}, Table~\ref{tab:accuracy_vector-based} highlights that all models perform substantially worse on visual tasks compared to other perceptual tasks. However, for humans, the difficulty gap between visual (85.20\%) and non-visual (88.00\%) tasks is minimal. This comparison suggests that while the questions themselves have comparable difficulty levels for humans, models struggle disproportionately with certain sensory dimensions. The best model performance on visual tasks (61.05\%) falls significantly short of human performance (85.20\%), indicating specific deficiencies in the models' embodied knowledge across different sensory domains rather than an issue of inherent question difficulty. Additionally, we analyzed the influence of potential confounding variables (see Appendix~\ref{sec:Confound variables}). In the Discussion section, we further examine performance variations across different subtasks.

\section{Discussion}
\label{tab:Discussion}

Experimental results on both datasets demonstrate that existing models perform poorly in representing and applying embodied knowledge, with no clear or systematic improvements observed in VLMs over text-only LMs. We further analyzed the models' performance across different question categories to investigate the specific aspects where models struggle most.

As shown in Figure~\ref{fig:box1}, on \texttt{SensoryVec}, models perform poorly on tactile and gustatory categories in the non-visual domain. In the visual domain, where overall performance is relatively weaker, models particularly struggle with color and shape. Table~\ref{tab:Models' Vector-Based Prediction Failures} shows some failure examples.\footnote{CLIP answers half correctly; GloVe, Word2Vec, and Qwen-7B each solve one; others fail all.}
It can be seen that sensory words judged more similar to their antonyms often exhibit greater similarity in form (e.g., \textit{sugarless-sugary}, \textit{unwrinkled-wrinkled}, \textit{undimmed-dim}) or frequency (e.g., \textit{dry-wet}, \textit{sharp-blunt}, \textit{open-closed}, \textit{green-red}), highlighting the limitations of distributed semantic representations in capturing semantic similarity.
% Vision-language models continue to heavily rely on distributional statistics, similar to unimodal models~\cite{ma-etal-2023-world}, which explains why they fail to achieve better performance on \texttt{SensoryVec}. Furthermore, in terms of error categories, tactile, visual, and gustatory domains exhibit the highest average concreteness ratings in the dataset, as shown in the Appendix~\ref{sec:Concreteness}. Concreteness refers to the degree to which words have a perceptible physical referent or are associated with tangible objects or experiences~\cite{ma-etal-2023-world}. This pattern suggests that categories where models perform worst tend to involve words with higher degrees of embodiment, further indicating that understanding meanings closely tied to embodied perception requires grounding beyond what distributed semantic representations can provide.

\begin{table*}[t]
\centering
\small  % 设置小字体
\resizebox{1\textwidth}{!}{
    \begin{tabular}{p{1\textwidth}}
    \toprule
    \textbf{Question \& Options \& Answers and Rationales from Claude3.5-Sonnet}\\
    \midrule
    \textbf{[V-GT]}~~If a rectangle has long sides on the top and bottom, and short sides on the left and right, and it is flipped around its left short side, where is the original bottom edge now positioned?  \\
    \textbf{A. Bottom}\quad B. Top\quad C. Right\quad zhe geD. Left  \\
    \emoji~~B. When a rectangle is flipped around its left short side (like a door hinge), the bottom edge will rotate 180 degrees upward, ending up at the top position. This is similar to how a door swings open, with the hinge being the left side in this case.\\
    \midrule
    \textbf{[V-S]}~~If the number 3 is horizontally cut in half, which direction does the opening of the bottom part face?  \\
    A. Up\quad \textbf{B. Left}\quad C. Down\quad D. Right  \\
    \emoji~~A. When a number 3 is cut horizontally in half, the bottom part resembles a curved line with the opening facing upward, similar to the bottom half of a sideways heart shape.\\
    \midrule
    \textbf{[V-B]}~~During the lower limb whipping motion, what is the trend of distance change between the knee and the hip?  \\
    A. Decreases\quad \textbf{B. Remains essentially unchanged}\quad C. Increases\quad D. Decreases first, then increases\quad  \\
    \emoji~~A. During a lower limb whipping motion, the knee moves closer to the hip as the leg flexes through the whipping action, resulting in a decrease in the distance between these two joints.\\
    \bottomrule
    \end{tabular}
}
% \vspace{-1.5mm}
\caption{Examples of models' \texttt{PerceptualQA} prediction failures and answers from Claude3.5-Sonnet. Task types are shown in brackets. Correct answers in \textbf{bold}.}
\label{tab:Models' QA-Based Prediction Failures}
\vspace{-2mm}
\end{table*}

% 实际应当放在conclusion前
% 微调：4o-mini提升较多，但视觉依然不如人类；qwen无提升，可能微调数据中答案不含选项文本，此处不提【提升较多是否要多提一下数据集的贡献？目前没提！】
\begin{table*}[t]
    \centering
    \resizebox{1\textwidth}{!}{
        \begin{tabular}{lllllllllllll}
        \toprule
        & \textbf{All Sensory} & \textbf{Visual} & \textbf{V-CA} & \textbf{V-CN} & \textbf{V-GT} & \textbf{V-S} & \textbf{V-B} & \textbf{Non-Visual} & \textbf{A} & \textbf{T} & \textbf{G} & \textbf{O}     \\
        \midrule
        Human Baseline & 86.00 & 85.20 & 66.00 & 88.00 & 89.00 & 93.00 & 90.00 & 88.00 & 88.00 & 88.00 & 86.00 & 90.00    \\
        GPT-4o-Mini & 58.21 & 50.96  & 57.14  & 71.43 & 42.86 & 37.21 & 47.83 & 79.17 & 75.00 & 80.00 & 92.86 & 72.22 \\
        GPT-4o-Mini-FT & 79.29 & 75.96 & 100.00 & 83.33 & 71.43 & 65.12 & 65.22 & 88.89      & 85.00 & 95.00 & 92.86 & 83.33 \\
        \bottomrule
        \end{tabular}
}
    % \vspace{-1.5mm}
    \caption{Accuracy of GPT-4o-Mini on \texttt{PerceptualQA} subtasks before and after fine-tuning.}
    \label{tab:gpt4o-mini-fine-tuning}
    \vspace{-5mm}
\end{table*}

% hu: The full task names are Visual-Color Attributes(V-CA), Visual-Colors in Nature(V-CN), Visual-Geometry and Transformations(V-GT), Visual-Symbols(V-S), Visual-Body(V-B), Auditory(A), Tactile(T), Gustatory(G), and Olfactory(O).
% 实际应当放在conclusion前

On \texttt{PerceptualQA} (Figure~\ref{fig:box2}), the symbol task poses the greatest challenge, followed by the body task, and the geometry and transformations task, while color-related tasks show comparatively better performance. 
We further examined the questions in the three most challenging visual understanding subtasks and list several difficult examples that no model could answer correctly in Table~\ref{tab:Models' QA-Based Prediction Failures}.\footnote{Results reflect only the models' first attempt out of two. Additional examples are provided in Appendix~\ref{sec:more_examples}.} We observed that models' errors are not tied to specific shapes (e.g., triangles), symbol types (e.g., numbers), body parts (e.g., arms), transformation types (e.g., rotation), or question targets (e.g., positional relationships). This observation suggests a systemic deficiency in spatial reasoning, rather than a lack of specific visual or conceptual knowledge.

Additionally, a potential explanation for why some visual tasks are more difficult is that reasoning about an object's transformed states often requires more complex inference than reasoning over comparative attributes of multiple entities. The design intention behind \texttt{PerceptualQA} is that both visual and non-visual questions require reasoning over a set of properties, as opposed to relying on shallow lexical associations. Specifically, when a question involves only a single target concept, we introduce transformations to elicit reasoning over derived attributes. For example, in the geometry and transformations task, we ask what shape is formed when a rectangle is folded along the line connecting the midpoints of its shorter sides. This requires understanding the geometric transformation of folding, as well as the shape of the object before and after the transformation. When the target concept involves multiple entities, the attribute set arises more directly. For instance, in the gustatory task, we ask how a set of foods should be grouped based on their taste. This involves identifying and comparing gustatory attributes and performing categorization. 

To investigate whether fine-tuning could directly improve models' understanding of embodied knowledge, we utilized the \texttt{PerceptualQA} dataset to fine-tune GPT-4o-Mini and Qwen2 series models. The results of GPT-4o-Mini are presented in Table~\ref{tab:gpt4o-mini-fine-tuning}. After supervised fine-tuning (SFT), the performance of GPT-4o-Mini improved from 58.21\% to 79.29\%, but still lagged behind human performance (86.00\%). This gap is primarily observed in the visual dimension, particularly in the three most challenging subtasks (V-GT, V-S, V-B) related to spatial perception and reasoning, where the fine-tuned model achieves an average accuracy of 67.26\% compared to 90.67\% of human performance. In addition, no significant performance improvements were observed in the Qwen2 series models. These results indicate that understanding embodied knowledge, especially spatial perception knowledge, remains a fundamental challenge for current models. Detailed experimental settings and results are presented in Appendix~\ref{sec:Fine-Tuning on QA-Based Dataset}.

% 不足的原因分析：训练机制
% 融合当前LLMs/VLLMs对于具身AI不足
VLLMs' limited improvement over LLMs in embodied knowledge understanding stems from their training mechanisms. 
% Our questions are intentionally designed to require sensory experiences rather than factual knowledge, challenging models since answering these questions depend on embodied imagination and reasoning that is rarely expressed in natural language.
Our questions are intentionally designed to require sensory experience rather than factual knowledge, making them challenging for models because answering them relies on embodied imagination and reasoning rarely conveyed in natural language.
VLLMs are typically trained on static image-text pairs that capture surface-level visual features. Due to the high cost of grounding annotations, it remains impractical to comprehensively label the knowledge about the physical world~\citep{ma-etal-2023-world}. Consequently, even models like Qwen2-VL~\citep{wang2024qwen2vlenhancingvisionlanguagemodels}, which adopt multi-stage training to improve image-text alignment, do not necessarily achieve better embodied understanding.\footnote{The three stages include: (1) training the vision encoder on image-text pairs, (2) leveraging larger-scale fine-grained datasets to refine visual-textual alignment, and (3) instruction tuning with multimodal and text-based dialogue data.} While multimodal LLMs are increasingly central to embodied AI research, their current limitations in perceiving and reasoning about embodied knowledge pose significant challenges to their effectiveness as the cognitive core of embodied agents. Our experimental results and analysis suggest that existing training tasks and datasets may be insufficient for fostering meaningful advances in this area.

\section{Conclusion}
% In this paper, we investigated the impact of visual grounding on models' embodied knowledge understanding. We developed an embodied knowledge understanding benchmark based on the perceptual system framework from psychology, encompassing external senses (visual, auditory, tactile, gustatory, and olfactory) and interoception. The benchmark features two tasks \texttt{SensoryVec} and \texttt{PerceptualQA}—comprising over 1,700 questions to systematically evaluate current SOTA models.

In this paper, we investigate how visual grounding influences models' understanding of embodied knowledge. To this end, we introduce a benchmark grounded in the psychological framework of the perceptual system, covering external senses (visual, auditory, tactile, gustatory, olfactory) as well as interoception. Our benchmark comprises two tasks—\texttt{SensoryVec} and \texttt{PerceptualQA}—with over 1,700 questions designed to systematically evaluate the models.

Our findings reveal that existing models perform suboptimally in embodied knowledge understanding, with vision-language models showing no significant advantage over text-only models. This suggests that current visual grounding approaches do not effectively enhance embodied knowledge comprehension. Further analysis shows that models' vector representations are susceptible to surface form and frequency bias, and they struggle with spatial perception and reasoning tasks. These insights underscore the importance of continued advancement in embodied knowledge understanding for AGI development. 

Future research should explore developing diverse forms of training data, novel training tasks and architectures that better integrate multimodal perceptual information to advance embodied knowledge understanding.
Specifically, in addition to text and image data, several categories of multimodal information may need to be incorporated, including dynamic sensory data such as video sequences, perception-related brain neural data, and feedback signals from non-human sensors such as haptic sensors and motion capture systems. These diverse data modalities may be essential to fully support the acquisition of comprehensive embodied knowledge.
Furthermore, future models may require joint training with embodied agents on foundational tasks that involve interaction with simulated or real-world environments, which would enable them to actively explore, ground concepts through experiential learning, and acquire causal understanding beyond statistical correlations.

\section*{Limitations}
% One limitation of this study is that \texttt{SensoryVec} was only evaluated on open-source models due to constraints in model availability. Consequently, the performance of state-of-the-art closed-source models on this dataset remains unknown.

The primary objective of our research was to diagnose the current models' understanding of embodied knowledge, rather than proposing methods to directly enhance model performance on this task. Although fine-tuning GPT-4o-Mini on \texttt{PerceptualQA} dataset yielded improved results, we have not yet fully explored the potential of purely textual data. Future work will aim to expand the size and diversity of the dataset, investigating its capacity to further boost model performance. 

% Moreover, we plan to examine training methods and model architectures that can more effectively harness visual information to enrich the embodied knowledge of LMs. By pursuing these avenues, we aim to gain deeper insights into how to imbue models with a more robust and nuanced embodied understanding.

\section*{Ethics Statement}
% During the research process, no sensitive or private information was used. The benchmark construction was subject to careful quality control, and we took great care to avoid introducing biased data during the dataset creation. To promote transparency and benefit the broader community, the benchmark has been made publicly available for future comparative evaluations and further research, supporting the development of accountable and trustworthy AI. The results were reported transparently, without manipulation.
No sensitive or private information was used in this research. Benchmark construction underwent rigorous quality control, with careful measures taken to avoid data bias. To promote transparency and benefit the broader community, the benchmark has been made publicly available for future comparative evaluations and further research. All results are reported transparently and without manipulation.

\section*{Acknowledgment}
The authors would like to thank Dr. Wang Yin and his lab members for their valuable discussions and suggestions. This research was supported by the Tencent Basic Platform Technology Rhino-Bird Focused Research Program.

% 致谢：审查版不需要
% \section*{Acknowledgments}
% Acknowledgments.

% 参考文献
% 让没有引用的文章也出现在文献列表
\nocite{}
% Bibliography entries for the entire Anthology, followed by custom entries
% ACLbib下载链接：\url{https://aclweb.org/anthology/anthology.bib.gz}.
%\bibliography{anthology,custom}
% Custom bibliography entries only
\bibliography{custom}

\newpage

% 附录
\appendix

\section{Motivation Behind Task Design}
\label{sec:Motivation Behind Task Design}

\subsection{SensoryVec Task}
\label{sec:Motivation Behind SensoryVec Task Design}
For pretrained language models, word vectors are learned based on the Distributional Semantic Hypothesis: words that occur in similar contexts tend to have similar meanings~\citep{harris1954distributional}. Since text-only models rely exclusively on contextual co-occurrences, they struggle to distinguish antonyms without incorporating additional mechanisms, such as contrastive mapping~\citep{Samenko_2021} or semantic information from thesauri~\citep{li2020distant}.

Specifically, regarding antonyms, their frequent co-occurrence in similar contexts (e.g., "this is a big ball" versus "this is a small ball") naturally leads to similarity in vector space. Therefore, antonyms represent a special case where terms share numerous semantic properties while differing primarily along a single dimension. These key dimensions often involve sensory attributes like size (big–small) or temperature (cold–warm), which are precisely the types of distinctions that text-only models struggle to differentiate due to their fundamentally non-perceptual architecture.

Visual data potentially introduces sensory information such as color, shape, and size, which may help differentiate concepts that appear in similar linguistic contexts but represent opposite meanings.

By employing vector similarity, our approach provides a targeted evaluation of whether multimodal models can encode sensory knowledge that distinguishes antonyms, a form of knowledge that is largely inaccessible to text-only models. This task is motivated by the need to assess whether visual grounding can overcome the fundamental limitations of text-based representations, thereby offering a meaningful strategy for evaluating perceptual grounding in model representation spaces.

\subsection{PerceptualQA Task}
\label{sec:Motivation Behind PerceptualQA Task Design}
When designing questions for \texttt{SensoryVec} task, we aimed for the question-answer pairs to rely on human sensory experiences rather than factual background knowledge (e.g., "What color is an apple?"). Each question is intended to be easily answerable by humans through embodied imagination and reasoning, but is rarely expressed in natural language. This introduces significant challenges for both text-only and multimodal models as the answers cannot be directly learned from training corpora through statistical co-occurrence patterns.

\section{Data Construction Details}
\label{sec:Process used for dataset generation}

\subsection{SensoryVec Task}
\label{sec:SensoryVec task}

\subsubsection{Word Data Cleaning Principles}
\label{sec:Word Selection Criteria}

% 修改一下最后一句为linesXXX.
For data that satisfied both sensory categorization and scoring criteria across multiple datasets, we performed the following filtering steps:
\begin{enumerate}
    \item Removed duplicate words within the same sensory category.
    \item Excluded words lacking documented synonyms or antonyms.
    \item Removed words absent from the established word embedding vocabulary.
    \item Excluded words for which it was difficult to construct natural contextual sentences that align with the principles outlined in Section~\ref{sec:Sentence Data Construction Details}.
\end{enumerate}

\subsubsection{Sentence Data Construction}
\label{sec:Sentence Data Construction Details}

Following the establishment of sensory triples (e.g., small–little–big), we curated three naturalistic contextual sentences for each triple according to the following procedure:

\begin{enumerate}
    \item First, we prompted GPT-4o to generate candidate sentences based on three criteria: (1) grammatical correctness, (2) semantic appropriateness, and (3) contextual alignment, which refers to the correspondence between each polysemous word’s meaning and its primary sensory modality.
    % 原来正文中内容：adhering to the following principles: (1) grammaticality, (2) semantic appropriateness, and (3) contextual alignment, matching polysemous words' meanings to their primary sensory category
    
    For sensory triples across diverse modalities, modality-specific prompts were developed. The following exemplifies a prompt employed for candidate generation:
    
    \begin{tcolorbox}[breakable, colback=myblue!50!white, colframe=myblue!60!black]
    Make English sentences so that any of the three adjectives \{small, little, big\} in the blank are grammatically correct and reasonable sentences. In addition, contextual information makes the meaning of these three words express what is perceived by the human eye. Give me ten sentences with spaces and no other information.
    \end{tcolorbox}

    \item Then, we reviewed and modified the generated sentences according to the aforementioned criteria. In instances where a sensory triple lacked sufficient valid sentences, supplementary sentences were manually constructed adhering to identical criteria.
\end{enumerate}

\subsection{PerceptualQA Task}
\label{sec:PerceptualQA task}

\subsubsection{Question Frameworks}
\label{sec:Question Frameworks}

For the \texttt{PerceptualQA} task, we adopt a high-level question design framework that formulates questions from multiple perspectives targeting a specific sensory dimension of the concept being examined. Each subtask follows its own question construction methodology tailored to the characteristics of that sensory domain. Comprehensive question frameworks are presented in Table~\ref{tab:question_frameworks_resized}.

\begin{table*}[ht]
\centering
\resizebox{1\textwidth}{!}{ % 调整宽度，高度自动适应
\begin{tabular}{p{2.5cm} p{8cm} p{6cm}} % 控制列宽，支持换行
\toprule
\textbf{Task} & \textbf{Target Concepts} & \textbf{Question Perspectives} \\
\midrule
Visual-Color Attributes & Objectively and subjectively described colors (e.g., red, cherry red) & Hue, brightness, saturation \\
\midrule
Visual-Colors in Nature & Edible animals (e.g., chicken), other animals (e.g., giraffe), ornamental plants (e.g., lucky bamboo), vegetables (e.g., carrot), fruits (e.g., banana), edible fungi (e.g., mushroom) & Hue, saturation, warm and cool tones \\
\midrule
Visual-Geometry and Transformations & Lines, triangles, quadrilaterals, polygons, circles, composite shapes; may undergo translation, rotation, flipping, compression, stretching, folding, cutting, combination & Quantity, shape, direction, size relation, position relation \\
\midrule
Visual-Symbols & Numbers (e.g., 6), letters (e.g., A), Chinese characters; may undergo translation, rotation, flipping, cutting, combination & Quantity, shape, direction, size relation, position relation \\
\midrule
Visual-Body & External body parts (e.g., head) in static postures or dynamic movements & Shape, distance, direction, speed, position relation \\
\midrule
Auditory & Sounds produced by objects or object interactions (e.g., baby crying, apple slicing) & Volume, pitch \\
\midrule
Tactile & Food, objects (e.g., shuttlecock), human body parts (e.g., neck) & Smoothness, elasticity, hardness, weight, thickness, stickiness, temperature \\
\midrule
Gustatory & Food & Sourness, sweetness, bitterness, saltiness, spiciness \\
\midrule
Olfactory & Food (vegetables, fruits, dishes, seasonings), other items (e.g., lucky bamboo), environments (e.g., hospital) & Fragrance, stench, specific odors \\
\bottomrule
\end{tabular}
}
\vspace{-2mm}
\caption{Question frameworks including target concepts and question perspectives in each \texttt{PerceptualQA} subtask.}
\vspace{-4mm}
\label{tab:question_frameworks_resized}
\end{table*}

\subsubsection{QA Data Construction}
\label{sec:QA Construction Details}

The construction of the \texttt{PerceptualQA} dataset adhered to the following procedures:
\begin{enumerate}
    \item We first prompted Claude-3.5-Sonnet as a brainstorming aid to generate candidate questions based on the question framework shown in Section~\ref{sec:Question Frameworks} to support human annotators. The majority of \texttt{PerceptualQA} items are manually crafted to meet rigorous design criteria, as most model-generated questions are discarded due to lacking a correct or unique answer, or being simply infeasible to answer. A small subset of questions originating from the model are manually screened and refined according to the same criteria. The design criteria are as follows:

    (1) Questions and corresponding answers should rely primarily on human sensory experience rather than factual background knowledge.
    
    (2) Questions should be readily answerable by humans through embodied imagination and reasoning, yet rarely expressed in natural language.
    
    (3) Each question should be objective, with a singular correct response among four provided options.
    
    (4) Incorrect options should be plausible, not obviously incorrect, thereby ensuring appropriate task difficulty.
    
    (5) All questions and answers should be clear, natural, and grammatically well-formed.

    For each task, multiple prompts were developed based on the established question framework. The following exemplifies a prompt utilized for candidate generation:

    \begin{tcolorbox}[breakable, colback=myblue!50!white, colframe=myblue!60!black]
    Construct 10 questions about the position relationships of body parts while lying.
    
    External body parts: head, shoulders, arms, hands, knees, feet, etc.
    
    Requirements: Answers should be objective choices. Return questions and options in JSON format.
    
    Example: \{"posture dimension": "lying", "question target": "position relationship", "question": "When lying flat, is the knee typically higher or lower than the hip when viewed from the side?", "options": \{"A": "Higher", "B": "Lower", "C": "Uncertain", "D": "Insufficient information to determine"\}\}
    
    Format: \{"posture dimension": "lying", "question target": "position relationship", "question": "", "options": \{"A": "", "B": "", "C": "", "D": ""\}\}
    \end{tcolorbox}

    \item Then, we reviewed and modified the generated questions according to the aforementioned criteria. In instances where the task exhibited insufficient question quantity, supplementary questions were manually formulated adhering to identical criteria.
\end{enumerate}

\section{Models and Experimental Settings Details}
\label{sec:model}
\subsection{SensoryVec Task}
We evaluated 16 models. For text-only models, we evaluated Word2Vec~\citep{mikolov2013efficient, mikolov2013distributed}, GloVe~\citep{pennington2014glove}, Bert-base-uncased (BERT)~\citep{devlin2018bert}, Qwen-7B~\citep{qwen}, Qwen2-7B~\citep{qwen2}, Qwen2-7B-Instruct~\citep{qwen2}, Mistral-7B-Instruct-v0.2 (Mistral-7B)~\citep{jiang2023mistral7b}, Vicuna-7b-v1.5 (Vicuna-7B)~\citep{zheng2023judging}, and GPT-2~\citep{radford2019language}. For VLMs, we evaluated VisualBERT~\citep{li2019visualbert}, Qwen-VL~\citep{Qwen-VL}, Qwen-VL-Chat~\citep{Qwen-VL}, Qwen2-VL-7B-Instruct~\citep{Qwen2VL, Qwen-VL}, LLaVA-v1.6-Mistral-7B (LLaVA1.6-Mistral-7B)~\citep{liu2024llavanext, liu2023improvedllava, liu2023llava}, LLaVA-v1.6-Vicuna-7B (LLaVA1.6-Vicuna-7B)~\citep{liu2024llavanext, liu2023improvedllava, liu2023llava}, and CLIP~\citep{radford2021learning}.

\subsection{PerceptualQA Task}
\label{sec:model_PerceptualQA Task}
We evaluated 24 models. For open-source LLMs, we evaluated Llama3.2-3b-instruct (Llama3.2-3B)~\citep{touvron2023llama}, Vicuna-7B, Mistral-7B, Qwen2-7B, Qwen2-7B-Instruct, Llama3.1-8b-instruct (Llama3.1-8B)~\citep{touvron2023llama}, Gemma2-9b-it (Gemma2-9B)~\citep{gemma_2024}, Gemma2-27b-it (Gemma2-27B)~\citep{gemma_2024}, Llama3.1-70b-instruct (Llama3.1-70B)~\citep{touvron2023llama}, Qwen2-72B-Instruct~\citep{qwen2}, and Llama3.1-405b-instruct (Llama3.1-405B)~\citep{touvron2023llama}. For closed-source LLMs, we evaluated OpenAI’s GPT-3.5-Turbo-0125 (GPT-3.5) and Alibaba Cloud's Qwen-Max-2024-09-19 (Qwen-Max). For open-source VLLMs, we evaluated LLaVA1.6-Vicuna-7B, LLaVA1.6-Mistral-7B, Qwen2-VL-7B-Instruct, and Qwen-VL-Max-2024-08-09 (Qwen2-VL-72B-Instruct)~\citep{Qwen2VL, Qwen-VL}. For closed-source VLMs, we evaluated Alibaba Cloud's Qwen-VL-Max-2024-11-19 (Qwen-VL-Max), Google’s Gemini-1.5-Flash-8B-001 (Gemini1.5-Flash-8B)~\citep{team2024gemini}, OpenAI’s GPT-4o-Mini-2024-07-18 (GPT-4o-Mini)~\citep{hurst2024gpt}, Google’s Gemini-1.5-Flash-002 (Gemini1.5-Flash)~\citep{team2024gemini}, OpenAI’s GPT-4o-2024-11-20 (GPT-4o)~\citep{hurst2024gpt}, Google’s Gemini-1.5-Pro-002 (Gemini1.5-Pro)~\citep{team2024gemini}, and Anthropic’s Claude-3.5-Sonnet-20241022 (Claude3.5-Sonnet).

In our evaluation, each question is asked independently with a corresponding prompt, eliminating the influence of historical context. The example of the prompt and question used in the evaluation is as follows:

\begin{tcolorbox}[breakable, colback=myblue!50!white, colframe=myblue!60!black]
\textbf{Prompt:}  
Based on the example provided, answer the question by selecting the most appropriate choice.  
Return your answer and rationale strictly in JSON format.

\medskip
\textbf{\#\#\#Example Input:}
\{
  "index": 000001,
  "question": "What color is the Fuji apple?",
  "options": \{
    "A": "Yellow",
    "B": "Green",
    "C": "Red",
    "D": "Blue"
  \}
\}

\textbf{\#\#\#Example Output:}
\{
  "index": 000001,
  "answer": "C",
  "rationale": "Different apple varieties come in different colors, and Fuji apples are typically red."
\}

\medskip
\textbf{\#\#\#Question:}
\{
  "index": 4064,
  "question": "How many triangles are there in the uppercase letter [A]?",
  "options": \{
    "A": "2",
    "B": "3",
    "C": "0",
    "D": "1"
  \}
\}

Return only the JSON.
\end{tcolorbox}
    
To ensure experiment reproducibility, the decoding temperature was set to 0.

To enhance the reliability of evaluation results and mitigate potential positional biases in model responses, we conducted two independent experiments for each model across the entire dataset. Each question was presented twice with randomly shuffled answer choices. Different random seeds were employed in each experiment to determine the position of the correct answer. The distribution of correct answers across these two experiments is presented in Table~\ref{tab:distribution_correct_answers}.

\begin{table}
  \centering
  \resizebox{0.45\textwidth}{!}{
      \begin{tabular}{lcccc}
        \toprule
        \textbf{Correct Option} & \textbf{A} & \textbf{B} & \textbf{C} & \textbf{D} \\
        \midrule
        \midrule
        Trial 1 & 338 & 371 & 325 & 366 \\
        Trial 2 & 323 & 353 & 357 & 367 \\
        \bottomrule
      \end{tabular}
  }
  \vspace{-2mm}
  \caption{Distribution of correct options across trials.}
  \label{tab:distribution_correct_answers}
  \vspace{-4mm}
\end{table}

% \section{Prompt for PerceptualQA}
% \label{sec:Prompt for QA-Based Dataset}
% In our evaluation, each question is asked independently with a corresponding prompt, eliminating the influence of historical context. The example of the prompt and question used in the evaluation is as follows:

% \begin{tcolorbox}[breakable, colback=myblue!50!white, colframe=myblue!60!black]
% \textbf{Prompt:}  
% Based on the example provided, answer the question by selecting the most appropriate choice.  
% Return your answer and rationale strictly in JSON format.

% \medskip
% \textbf{\#\#\#Example Input:}
% \{
%   "index": 000001,
%   "question": "What color is the Fuji apple?",
%   "options": \{
%     "A": "Yellow",
%     "B": "Green",
%     "C": "Red",
%     "D": "Blue"
%   \}
% \}

% \textbf{\#\#\#Example Output:}
% \{
%   "index": 000001,
%   "answer": "C",
%   "rationale": "Different apple varieties come in different colors, and Fuji apples are typically red."
% \}

% \medskip
% \textbf{\#\#\#Question:}
% \{
%   "index": 4064,
%   "question": "How many triangles are there in the uppercase letter [A]?",
%   "options": \{
%     "A": "2",
%     "B": "3",
%     "C": "0",
%     "D": "1"
%   \}
% \}

% Return only the JSON.
% \end{tcolorbox}

\section{Result Details}
\label{sec:Result Details}

\subsection{Fine-Tuning on PerceptualQA Dataset}
\label{sec:Fine-Tuning on QA-Based Dataset}

We further investigated the feasibility of directly utilizing the \texttt{PerceptualQA} dataset as training data for fine-tuning models, with the explicit aim of enhancing their embodied knowledge comprehension. Table~\ref{tab:fine-tuning} presents comparative accuracies before and after the fine-tuning process. 

% Additionally, the performance of GPT-4o-Mini on subtasks is shown in Table~\ref{tab:gpt4o-mini-fine-tuning}.

\vspace{-2mm}
\begin{table}[h]
    \centering
    \resizebox{0.45\textwidth}{!}{
        \begin{tabular}{llcc}
            \toprule
            Model & Method & Before & After \\
            \midrule
            GPT-4o-Mini & SFT & 58.21 & \textbf{79.29} \\
            \midrule
            Qwen2-7B & LoRA & 47.50 & 45.00 \\
            Qwen2-7B-Instruct & LoRA & 46.43 & 46.43 \\
            Qwen2-VL-7B & LoRA & 47.86 & 48.21 \\
            \bottomrule
        \end{tabular}
    }
    \vspace{-2mm}
    \caption{Accuracy of GPT-4o-Mini and Qwen2 series models on \texttt{PerceptualQA} before and after fine-tuning.}
    \label{tab:fine-tuning}
    \vspace{-2mm}
\end{table}

For our experimental design, we randomly shuffled the \texttt{PerceptualQA} dataset and allocated 80\% for training and the remaining 20\% for testing. The training data comprised 1,120 question-answer pairs, stored in JSON format. Notably, since the Qwen2-7B series models demonstrated suboptimal performance in generating structured outputs after fine-tuning, we configured these models to output only the correct answer option. In contrast, for GPT-4o-Mini, we implemented a more comprehensive output format including both the correct option and its specific content. This distinction in output requirements may partially account for performance differentials between models. We conducted supervised fine-tuning (SFT) on the GPT-4o-Mini model, while employing Low-Rank Adaptation (LoRA) tuning~\citep{hu2021lora} for the Qwen2 series models due to computational resource constraints.

The precise experimental configurations were as follows: The GPT-4o-Mini model underwent fine-tuning for 3 epochs with a batch size of 2 and a learning rate multiplier of 1.8. To ensure reproducibility, we established a seed value of 656770515 for the fine-tuning procedure. We fine-tuned three models from the Qwen2 series using the LLaMA Factory framework~\citep{zheng2024llamafactory}: Qwen2-7B, Qwen2-7B-Instruct, and Qwen2-VL-7B-Instruct. We employ the Low-Rank Adaptation (LoRA)~\citep{hu2021lora} method for parameter-efficient fine-tuning, targeting the query and value projection matrices in the attention layers. For Qwen2-7B, we implemented the Alpaca template for instruction tuning, while Qwen2-7B-Instruct and Qwen2-VL-7B-Instruct utilized the Qwen and Qwen2-VL templates, respectively. All models underwent fine-tuning for 3 epochs with a batch size of 4 and a learning rate of 5e-6, utilizing the cosine learning rate scheduler. We applied gradient accumulation every 4 steps and conducted evaluations at 1,000-step intervals. The maximum sequence length was configured at 4,096 tokens, with weight decay set to 0.1 and a warm-up period of 100 steps. Training was conducted using 16-bit bfloat16 precision, with each model allocated to a separate NVIDIA GeForce RTX 4090 GPU.

Fine-tuning data examples for both GPT-4o-Mini and Qwen2 series models are presented herein. During the inference phase, we employed identical data formatting as used in the fine-tuning process, with the "answer" field information removed to serve as the prompt.

Fine-tuning Data Sample for GPT-4o-Mini:
\begin{tcolorbox}[breakable, colback=myblue!50!white, colframe=myblue!60!black]
\{"messages": [\{"role": "system", "content": "Based on the example provided, please answer the question by selecting the most appropriate choice. Provide your answer in JSON format, and return only the JSON.\#\#\#Example Input\{"question": "What color is the Fuji apple?","options": \{"A": "Yellow","B": "Green","C": "Red","D": "Blue"\}\}\#\#\#Example Output\{"answer": "C.Red"\}\#\#\#Question"\}, \{"role": "user", "content": "\{"question": "How many triangles are there in the uppercase letter [A]?", "options": \{"A": "2", "B": "3", "C": "0", "D": "1"\}\}Return only the JSON."\}, \{"role": "assistant", "content": "\{"answer": "D.1"\}"\}]\}
\end{tcolorbox}

Fine-tuning Data Sample for Qwen2 series:
\begin{tcolorbox}[breakable, colback=myblue!50!white, colframe=myblue!60!black]
\{"instruction": "Based on the example provided, please answer the question by selecting the most appropriate choice. Provide your answer in JSON format, and return only the JSON.\#\#\#Example Input\{"question": "What color is the Fuji apple?","options": \{"A": "Yellow","B": "Green","C": "Red","D": "Blue"\}\}\#\#\#Example Output\{"answer": "C"\}\#\#\#Question\{"question": "How many triangles are there in the uppercase letter [A]?", "options": \{"A": "2", "B": "3", "C": "0", "D": "1"\}\}Return only the JSON.", "input": "", "output": "D"\}
\end{tcolorbox}

\subsection{Cross-Language Generalization on PerceptualQA Task}
\label{sec:Cross-Language Generalization}
\begin{table*}[htbp]
  \centering
  \resizebox{1\textwidth}{!}{
    \begin{tabular}{lcccccccccccc}
      \toprule
      \textbf{Model} & \textbf{All Sensory} & \textbf{Visual} & \textbf{V-CA} & \textbf{V-CN} & \textbf{V-GT} & \textbf{V-S} & \textbf{V-B} & \textbf{Non-Visual} & \textbf{A} & \textbf{T} & \textbf{G} & \textbf{O}\\
      \midrule
      \midrule
      \multicolumn{13}{c}{\textbf{English}} \\
      \midrule
      Qwen2-7B & 49.36 & 39.85 & 42.00 & 58.75 & 35.00 & 28.75 & 34.75 & 73.13 & 68.50 & 62.00 & 83.00 & 79.00\\
      Qwen2-7B-Instruct & 47.82 & 35.80 & 41.00 & 59.75 & 26.75 & 22.00 & 29.50 & 77.88 & 73.50 & 68.00 & 82.50 & 87.50\\
      Qwen2-VL-7B-Instruct & 51.00 & 41.70 & 43.75 & 59.75 & 38.25 & 31.00 & 35.75 & 74.25 & 69.50 & 64.00 & 86.00 & 77.50\\
      \midrule
      \multicolumn{13}{c}{\textbf{Chinese}} \\
      \midrule
      Qwen2-7B & 51.72 & 41.95 & 47.50 & 64.75 & 34.75 & 22.50 & 40.25 & 76.13 & 62.50 & 70.50 & 87.00 & 84.50\\
      Qwen2-7B-Instruct & 51.54 & 39.90 & 46.50 & 60.00 & 33.25 & 24.00 & 35.75 & 80.63 & 72.50 & 73.00 & 86.50 & 90.50\\
      Qwen2-VL-7B-Instruct & 52.54 & 41.65 & 52.75 & 58.00 & 37.50 & 25.75 & 34.25 & 79.75 & 71.00 & 71.50 & 88.50 & 88.00\\
      \bottomrule
      \end{tabular}
  }
  \vspace{-2mm}
  \caption{Accuracy of \texttt{PerceptualQA} in English and Chinese.The full task names are Visual-Color Attributes(V-CA), Visual-Colors in Nature(V-CN), Visual-Geometry and Transformations(V-GT), Visual-Symbols(V-S), Visual-Body(V-B), Auditory(A), Tactile(T), Gustatory(G), and Olfactory(O).}
  \label{tab:accuracy_QA-based_in_Chinese}
  \vspace{-2mm}

\end{table*}

% 补充：比较中英文表现差异
In order to validate the cross-language generalizability of our findings, we assessed the performance of the Qwen2 series models on the Chinese \texttt{PerceptualQA}. As illustrated in Table~\ref{tab:accuracy_QA-based_in_Chinese}, the results indicate that VLMs do not exhibit a significant improvement compared to the text-only models. Furthermore, comparative analysis of model performance across both Chinese and English versions of \texttt{PerceptualQA} reveals consistent overall performance, confirming the reliability of our evaluation approach across these two languages. Both English and Chinese versions of the dataset are made available.

% The original dataset was created in Chinese and translated using Claude3.5-Sonnet, followed by manual proofreading. 

\subsection{Potential Confounding Variables in PerceptualQA Task}
\label{sec:Confound variables}
We analyzed all questions in the \texttt{PerceptualQA} dataset for length, average word frequency, and average Age of Acquisition (AoA) scores, and then conducted several analyses.

\subsubsection{Method}
We computed question length based on token count including all options. Word frequency was calculated using the \texttt{wordfreq} Python library. Average AoA scores were derived for each question and its four answer options using \citeposs{kuperman2012age} established ratings, excluding words absent from the Kuperman dataset.

\subsubsection{Results across tasks}
We calculated the mean and standard deviation of three metrics for each task. Our analysis revealed that the V-S, V-GT, and V-B questions, which demonstrated the lowest performance, did not exhibit significantly higher or lower values in sentence length or AoA scores. However, these question types displayed slightly higher word frequencies compared to other types. This likely stems from their focus on basic geometric shapes, numbers, and body-related nouns, in contrast to other types that involved animal and plant names. Based on these three linguistic metrics, we concluded that question type itself exerts a more substantial impact on model performance than these potential confounding variables.

\begin{table}[ht]
    \centering
    \resizebox{0.45\textwidth}{!}{
        \begin{tabular}{llcc}
            \toprule
            \textbf{Task} & \textbf{Length} & \textbf{Average word frequency} & \textbf{Average AoA scores} \\
            \midrule
            V-CA & 18.50 (2.38) & 5.34 (0.17) & 5.55 (0.71) \\
            V-CN & 54.58 (24.87) & 4.76 (0.22) & 6.07 (0.75) \\
            V-GT & 33.45 (11.74) & 5.50 (0.31) & 5.77 (0.58) \\
            V-S & 23.02 (7.45) & 5.60 (0.31) & 5.30 (0.45) \\
            V-B & 30.23 (5.54) & 5.70 (0.18) & 5.25 (0.40) \\
            A & 30.29 (10.62) & 5.47 (0.32) & 4.79 (0.41) \\
            T & 13.96 (3.84) & 4.85 (0.29) & 5.81 (0.85) \\
            G & 26.75 (19.20) & 5.22 (0.20) & 5.29 (0.52) \\
            O & 16.84 (5.43) & 5.07 (0.38) & 5.89 (0.64) \\
            \bottomrule
        \end{tabular}
    }
    \vspace{-2mm}
    \caption{Statistical properties of questions across tasks. Values are presented as mean (standard deviation).}
    \label{tab:question_stats}
    \vspace{-5mm}
\end{table}

\subsubsection{Comparison between correctly and incorrectly answered questions}
We further conducted statistical analyses comparing questions answered correctly versus incorrectly by 24 models.

Regarding question length, most models showed similar length distributions between both question sets, with typical differences of only 1-3 words. However, Mann-Whitney U tests revealed incorrectly answered questions were significantly longer. For most models, incorrectly answered questions had significantly higher average word frequencies. No significant differences in AoA metrics were observed between the two question sets for most models. This analysis suggests that question length and average word frequency may influence model performance if we don't take task type into consideration. 

\subsection{Detailed Results on PerceptualQA Task}
\label{sec:Detailed Results on PerceptualQA}
We present the results of all models across each subtask, with detailed results shown in Table~\ref{tab:detailed_accuracy_QA-based}.

\begin{table*}[ht]
  \centering
  \resizebox{0.95\textwidth}{!}{
    \begin{tabular}{lllllllllllll}
    \toprule
    \textbf{Model}                      & \textbf{All Sensory} & \textbf{Visual}  & \textbf{V-CA}    & \textbf{V-CN}    & \textbf{V-GT}    & \textbf{V-S}     & \textbf{V-B}     & \textbf{Non-Visual} & \textbf{A}       & \textbf{T}       & \textbf{G}       & \textbf{O}   \\
    \midrule
    Human Baseline & 86.00 & 85.20 & 66.00 & 88.00 & 89.00 & 93.00 & 90.00 & 88.00 & 88.00 & 88.00 & 86.00 & 90.00    \\
    \midrule
    Llama3.2-3B-Instruct       & 47.71     & 38.25 & 47.50 & 46.75 & 31.75 & 27.75 & 37.50 & 71.38    & 67.00 & 61.50 & 82.00 & 75.00  \\
    Vicuna-7B             & 38.25     & 32.60 & 30.50 & 33.00 & 35.25 & 30.25 & 34.00 & 52.38    & 48.50 & 38.00 & 65.00 & 58.00  \\
    LlaVa1.6-Vicuna-7B       & 41.64     & 35.25 & 37.50 & 37.25 & 30.50 & 31.75 & 39.25 & 57.63    & 49.50 & 49.00 & 70.50 & 61.50  \\
    Mistral-7B   & 42.96     & 31.40 & 42.50 & 35.50 & 26.25 & 24.25 & 28.50 & 71.88    & 67.00 & 65.50 & 78.50 & 76.50  \\
    Llava1.6-Mistral-7B      & 45.64     & 35.90 & 45.00 & 40.00 & 30.75 & 25.00 & 38.75 & 70.00    & 71.00 & 66.50 & 70.50 & 72.00  \\
    Qwen2-7B                   & 49.36     & 39.85 & 42.00 & 58.75 & 35.00 & 28.75 & 34.75 & 73.13    & 68.50 & 62.00 & 83.00 & 79.00  \\
    Qwen2-7B-Instruct          & 47.82     & 35.80 & 41.00 & 59.75 & 26.75 & 22.00 & 29.50 & 77.88    & 73.50 & 68.00 & 82.50 & 87.50  \\
    Qwen2-VL-7B-Instruct       & 51.00     & 41.70 & 43.75 & 59.75 & 38.25 & 31.00 & 35.75 & 74.25    & 69.50 & 64.00 & 86.00 & 77.50  \\
    Llama3.1-8B       & 48.54     & 39.05 & 46.00 & 49.25 & 33.50 & 27.50 & 39.00 & 72.25    & 70.50 & 61.00 & 82.00 & 75.50  \\
    Gemma2-9B              & 51.89     & 40.65 & 45.50 & 55.50 & 38.00 & 28.25 & 36.00 & 80.00    & 81.50 & 72.50 & 83.00 & 83.00  \\
    Gemma2-27B             & 55.39     & 44.90 & 51.50 & 61.25 & 43.75 & 30.75 & 37.25 & 81.63    & 81.50 & 73.50 & 85.00 & 86.50  \\
    Llama3.1-70B      & 59.71     & 49.85 & 65.50 & 63.25 & 47.75 & 32.25 & 40.50 & 84.38    & 84.00 & 77.50 & 89.00 & 87.00  \\
    Qwen2-72B-Instruct         & 62.32     & 52.75 & 69.75 & 69.75 & 45.25 & 35.00 & 44.00 & 86.25    & 84.50 & 84.00 & 88.00 & 88.50  \\
    Qwen2-VL-72B-Instruct     & 63.89     & 54.45 & 68.75 & 73.00 & 51.00 & 32.75 & 46.75 & 87.50    & 83.00 & 86.50 & 89.50 & 91.00  \\
    Llama3.1-405B     & 63.46     & 54.55 & 69.75 & 72.00 & 47.50 & 37.00 & 46.50 & 85.75    & 83.50 & 80.50 & 90.00 & 89.00  \\
    GPT-3.5         & 50.46     & 39.50 & 40.50 & 42.50 & 43.75 & 29.25 & 41.50 & 77.88    & 73.00 & 69.50 & 84.50 & 84.50  \\
    Qwen-Max        & 68.71     & 61.05 & 85.00 & 74.50 & 57.25 & 44.25 & 44.25 & 87.88    & 87.00 & 84.50 & 91.00 & 89.00  \\
    GPT-4o-Mini     & 57.18     & 46.35 & 57.25 & 62.25 & 41.25 & 31.50 & 39.50 & 84.25    & 84.50 & 78.50 & 86.00 & 88.00  \\
    Gemini1.5-Flash-8B        & 54.39     & 44.55 & 56.75 & 59.50 & 42.50 & 28.50 & 35.50 & 79.00    & 82.00 & 67.50 & 85.00 & 81.50  \\
    Gemini1.5-Flash       & 56.07     & 45.20 & 65.75 & 48.25 & 42.75 & 31.25 & 38.00 & 83.25    & 81.00 & 76.50 & 88.50 & 87.00  \\
    GPT-4o          & 68.46     & 59.45 & 70.00 & 77.50 & 51.25 & 44.50 & 54.00 & 91.00    & 92.50 & 86.00 & 91.00 & 94.50  \\
    Gemini1.5-Pro         & 65.21     & 56.55 & 70.00 & 70.50 & 54.00 & 39.00 & 49.25 & 86.88    & 82.50 & 83.50 & 89.50 & 92.00  \\
    Claude3.5-Sonnet & 69.04     & 60.00 & 70.25 & 75.25 & 57.00 & 44.75 & 52.75 & 91.63    & 92.00 & 90.00 & 92.00 & 92.50  \\
    Qwen-VL-Max     & 64.68     & 55.30 & 68.75 & 79.25 & 50.00 & 34.25 & 44.25 & 88.13    & 84.50 & 87.00 & 90.50 & 90.50  \\
    \bottomrule
    \end{tabular}
  }
  \vspace{-2mm}
  \caption{Detailed Results on \texttt{PerceptualQA}. The full task names are Visual-Color Attributes(V-CA), Visual-Colors in Nature(V-CN), Visual-Geometry and Transformations(V-GT), Visual-Symbols(V-S), Visual-Body(V-B), Auditory(A), Tactile(T), Gustatory(G), and Olfactory(O).}
  \label{tab:detailed_accuracy_QA-based}
  \vspace{-3mm}
\end{table*}

% \section{Concreteness Scores Across Sensory Categories in SensoryVec}
% \label{sec:Concreteness}
% We present the results of the concreteness analysis across sensory categories in \texttt{SensoryVec} in Table~\ref{tab:concreteness_scores}. Specifically, we calculated the average concreteness score for all words within each sensory category based on the ratings provided by~\citet{brysbaert2014concreteness}. Words not covered in their lexicon were excluded from the analysis.

% \begin{table}[htbp]
%   \centering
%   \resizebox{0.45\textwidth}{!}{
%     \begin{tabular}{l c}
%     \toprule
%     \textbf{Sensory Category} & \textbf{Average Concreteness Score} \\
%     \midrule
%     Visual     & 3.34 \\
%     Auditory   & 2.93 \\
%     Tactile & 3.57 \\
%     Gustatory  & 3.09 \\
%     Olfactory  & 3.07 \\
%     Interoceptive & 2.48 \\
%     \bottomrule
%     \end{tabular}
%   }
%   \caption{Average concreteness scores across sensory categories in \texttt{SensoryVec}.}
%   \label{tab:concreteness_scores}
% \end{table}

\subsection{More Examples of models’ PerceptualQA prediction failures}
\label{sec:more_examples}
We present additional examples of models’ \texttt{PerceptualQA} prediction failures in Table~\ref{tab:More Models' QA-Based Prediction Failures}.

\begin{table*}[ht]
\centering
\small  % 设置小字体
\resizebox{0.95\textwidth}{!}{
    \begin{tabular}{p{1\textwidth}}
    \toprule
    \textbf{Question \& Options \& Answers and Rationales from Claude3.5-Sonnet}\\
    \midrule
    \textbf{[V-GT]}~~If a rectangle has long sides on the top and bottom, and short sides on the left and right, and it is flipped around its left short side, where is the original bottom edge now positioned?  \\
    \textbf{A. Bottom}\quad B. Top\quad C. Right\quad D. Left  \\
    \emoji~~B. When a rectangle is flipped around its left short side (like a door hinge), the bottom edge will rotate 180 degrees upward, ending up at the top position. This is similar to how a door swings open, with the hinge being the left side in this case.\\
    \midrule
    \textbf{[V-S]}~~If the number 3 is horizontally cut in half, which direction does the opening of the bottom part face?  \\
    A. Up\quad \textbf{B. Left}\quad C. Down\quad D. Right  \\
    \emoji~~A. When a number 3 is cut horizontally in half, the bottom part resembles a curved line with the opening facing upward, similar to the bottom half of a sideways heart shape.\\
    \midrule
    \textbf{[V-B]}~~When standing and bending one leg backward at the knee, which is higher: the ankle of the raised leg or the knee of the supporting leg?  \\
    \textbf{A. It depends}\quad B. They are at the same height\quad C. The knee of the supporting leg\quad D. The ankle of the raised leg  \\
    \emoji~~D. When bending one leg backward at the knee while standing, the ankle of the raised leg typically rises above the knee level of the supporting leg. This is due to the natural biomechanics of the leg and the way the knee joint bends, allowing the foot and ankle to be lifted higher than the stationary knee of the supporting leg.\\
    \bottomrule
    \end{tabular}
}
\vspace{-2mm}
\caption{More Examples of models' \texttt{PerceptualQA} prediction failures and answers from Claude3.5-Sonnet. Task types are shown in brackets. The correct answers are shown in \textbf{bold}.}
\label{tab:More Models' QA-Based Prediction Failures}
\vspace{-5.5mm}
\end{table*}

\subsection{Human Baseline Details for PerceptualQA Task}
\label{sec:Human Participant Information}

For \texttt{PerceptualQA}, we established a human baseline with native speakers to provide a reference for evaluating model performance.

Specifically, we recruited seven graduate students, with each question receiving responses from two different participants. We randomly selected 25\% of the questions from each of the nine tasks, shuffled them, and divided them into seven sets of 50 questions each, ensuring proportional representation of all nine subtasks across sets. Each participant completed two sets (100 questions total), following a fixed task order within each set. This design minimizes individual participant influence on task-level accuracy and establishes a reliable human performance benchmark. For each subtask, the human baseline was computed as the average accuracy of the two responses.

All participants reported good physical health with no relevant impairments (e.g., color blindness). Screening was conducted during recruitment based on self-reported information to confirm participants met the health criteria and could complete the assessment within the designated timeframe. Participants completed the tasks independently via an online survey platform. The use of LLM tools was prohibited. On average, participants required 37.32 minutes to complete the assigned questions. Recruitment occurred through announcements within graduate student communities at our institution. Participants were explicitly informed that their responses would be used for research purposes. No personally identifiable information was retained in association with participant responses. Each participant received compensation of 50 RMB, equivalent to 0.5 RMB per question.

\end{document}